\pdfoutput=1

\documentclass[11pt]{article}

\usepackage{acl}

\usepackage{times}
\usepackage{latexsym}
\usepackage{url}
\usepackage{hyperref}
\usepackage{diagbox}
\makeatletter
\def\UrlAlphabet{%
      \do\a\do\b\do\c\do\d\do\e\do\f\do\g\do\h\do\i\do\j%
      \do\k\do\l\do\m\do\n\do\o\do\p\do\q\do\r\do\s\do\t%
      \do\u\do\v\do\w\do\x\do\y\do\z\do\A\do\B\do\C\do\D%
      \do\E\do\F\do\G\do\H\do\I\do\J\do\K\do\L\do\M\do\N%
      \do\O\do\P\do\Q\do\R\do\S\do\T\do\U\do\V\do\W\do\X%
      \do\Y\do\Z}
\def\UrlDigits{\do\1\do\2\do\3\do\4\do\5\do\6\do\7\do\8\do\9\do\0}
\g@addto@macro{\UrlBreaks}{\UrlOrds}
\g@addto@macro{\UrlBreaks}{\UrlAlphabet}
\g@addto@macro{\UrlBreaks}{\UrlDigits}
\makeatother

\usepackage[T1]{fontenc}

\usepackage[utf8]{inputenc}

\usepackage{microtype}
\usepackage{amsmath}
\usepackage{bbm}
\usepackage{colortbl}
\usepackage{soul, color, xcolor}
\definecolor{green-table}{rgb}{.8828125,.93359375,.8515625}
\definecolor{blue-table}{rgb}{.84765625,.87890625,.9453125}
\usepackage{array}
\usepackage{graphicx}      
\usepackage{booktabs}      
\usepackage{subcaption}    
\usepackage{multirow}      
\usepackage{multicol}      
\usepackage{amssymb}       
\usepackage{pifont}

\usepackage{comment}
\usepackage{enumerate}
\usepackage{enumitem}

\usepackage{inconsolata}

\title{Think Before You Act: A Two-Stage Framework for Mitigating Gender Bias Towards Vision-Language Tasks}

\author{Yunqi Zhang\textsuperscript{1}, Songda Li\textsuperscript{1}, Chunyuan Deng\textsuperscript{2}, Luyi Wang\textsuperscript{1}, Hui Zhao\textsuperscript{1,3,}\thanks{\ \  Corresponding author.}  \\
        \textsuperscript{1}Software Engineering Institute, East China Normal University  \\ 
        \textsuperscript{2}Georgia Institute of Technology \\
        \textsuperscript{3}Shanghai Key Laboratory of Trustworthy Computing, Shanghai, China \\ 
        \texttt{\{yunqi.zhang,songda.li,luyi.wang\}@stu.ecnu.edu.cn} \\
        \texttt{cdeng73@gatech.edu} \\
        \texttt{hzhao@sei.ecnu.edu.cn}
}

\begin{document}
\maketitle

\begin{abstract}

Gender bias in vision-language models (VLMs) can reinforce harmful stereotypes and discrimination.
In this paper, we focus on mitigating gender bias towards vision-language tasks. We identify object hallucination as the essence of gender bias in VLMs. Existing VLMs tend to focus on salient or familiar attributes in images but ignore contextualized nuances. Moreover, most VLMs rely on the co-occurrence between specific objects and gender attributes to infer the ignored features, ultimately resulting in gender bias. 
We propose GAMA, a task-agnostic \underline{g}eneration fr\underline{a}mework to \underline{m}itigate gender bi\underline{a}s. GAMA consists of two stages: narrative generation and answer inference. 
During narrative generation, GAMA yields all-sided but gender-obfuscated narratives, which prevents premature concentration on localized image features, especially gender attributes. 
During answer inference, GAMA integrates the image, generated narrative, and a task-specific question prompt to infer answers for different vision-language tasks. This approach allows the model to rethink gender attributes and answers.
We conduct extensive experiments on GAMA, demonstrating its debiasing and generalization ability.\footnote{Our code is available at \url{https://github.com/zyq0000/GAMA}.}

\end{abstract}

\section{Introduction}
\label{sec:intro}

\begin{figure}[t]
    \centering
    \includegraphics[width=\linewidth]{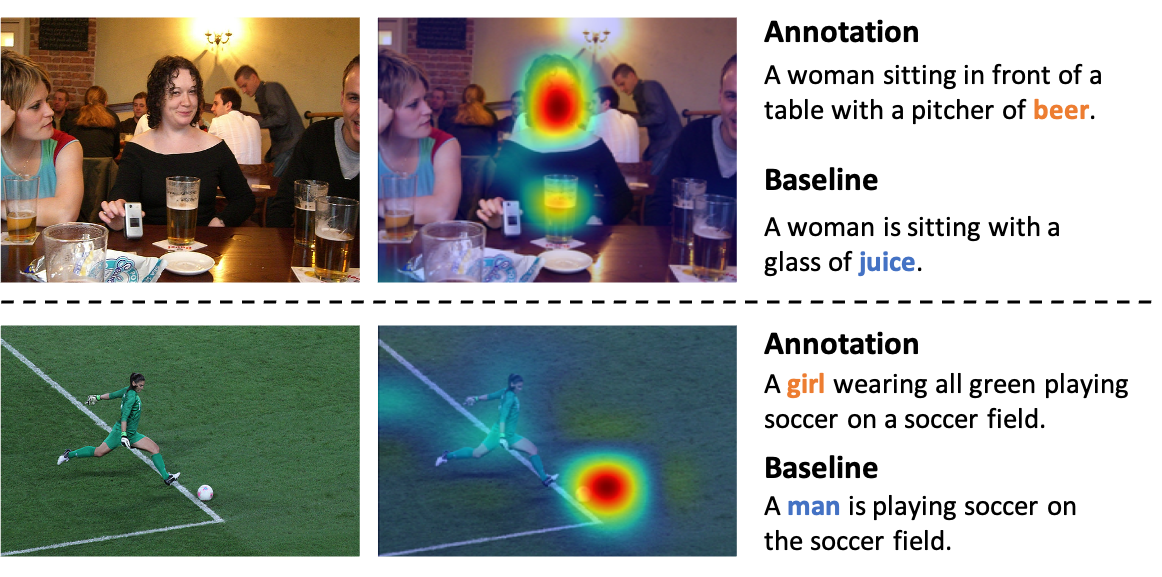}
    \caption{\textbf{Examples in image captioning with annotations and captions generated by a baseline model, SAT \citep{xu_show_2015}.} We overlay images with attention heatmaps of SAT on the right. In the top example, SAT focuses on the woman and predicts ``juice'', a word highly co-occurring with females. In the bottom example, the gender is incorrectly predicted, as ``soccer'' highly co-occurs with males in the training set.}
    \label{fig:intro-bias-examples}
\end{figure}

Vision-language models (VLMs) have attracted significant attention in recent years due to their widespread applications in image captioning \citep{li_oscar_2020,nguyen_grit_2022}, image-text retrieval \citep{wang_consensus-aware_2020,qu_dynamic_2021}, and visual question answering \citep{antol_vqa_2015,jiang_defense_2020}. 
Remarkable advancements have been achieved in these tasks, primarily measured by task performance metrics \citep{radford_learning_2021,li_blip-2_2023}.
However, there is a growing concern about the undesirable social bias (e.g., gender, race) in VLMs \citep{hendricks_women_2018,ross_measuring_2021,zhang_counterfactually_2022}. An example is presented in Figure~\ref{fig:intro-bias-examples} (bottom), implying stereotypes associating sports with masculinity.
More alarmingly, biased VLMs have the potential to propagate and even exacerbate existing stereotypes and inequalities.

Previous methods for mitigating social bias towards vision-language tasks can be grouped into two categories: 1) \textit{task-specific methods} are designed for task-specific datasets and models, which mitigate bias by re-sampling datasets \citep{zhao_men_2017,wang_are_2021}, synthesizing negative samples for training \citep{hirota_model-agnostic_2023}, or introducing debiasing modules or training objectives to task-specific models \citep{hendricks_women_2018,tang_mitigating_2021,seth_dear_2023}; and 2) \textit{task-agnostic methods} aim to pre-train a debiased encoder for downstream tasks, most commonly by adopting an adversarial approach to remove unwanted features \citep{wang_balanced_2019}, or leveraging counterfactual samples to minimize biased representations \citep{zhang_counterfactually_2022}. 
Despite these efforts, task-specific methods often lack generalization ability, while existing task-agnostic methods primarily address bias at the feature level and fall short of probing the essence of social bias.

In this work, we focus on gender bias as it is a crucial aspect of social bias. To address potential limitations in existing research, our initial step is to explore the essence of gender bias. 
We posit that gender bias is a manifestation of object hallucination in VLMs \citep{rohrbach_object_2018}. 
Specifically, models understand the form rather than the meaning.\footnote{Following \citet{bender_climbing_2020}, we define \textit{form} as the visible realization of vision and language, such as pixels or bytes in digital representations of text or image, and \textit{meaning} as the relationship between form and external elements to vision and language, like communicative intent.} 
This leads to a tendency for VLMs to focus on the most salient or familiar objects or attributes but ignore the rest of the image. 
Moreover, as depicted in Figure \ref{fig:intro-bias-examples}, VLMs are inclined to hallucinate objects that co-occur with gender words, and vice versa for gender attributes closely associated with specific objects in the training set.
As a result, VLMs may generate answers containing objects or gender attributes inconsistent with the given image.

We propose GAMA, a novel task-agnostic \underline{g}en-eration fr\underline{a}mework to \underline{m}itigate gender bi\underline{a}s towards vision-language tasks. Unlike previous methods, GAMA is a multi-level method, addressing bias both in external object co-occurrences and in internal bias features.
The framework comprises two stages: \textit{narrative generation} and \textit{answer inference}. During \textit{narrative generation}, GAMA creates an all-sided narrative for a given image, which prevents premature focus on localized details. Besides, we disentangle features through contrastive learning \citep{he_momentum_2020, chen_simple_2020} to obfuscate gender information in the generated narratives. This approach helps mitigate the influence of gender attributes on context generation. The trained model can seamlessly adapt to different vision-language tasks for narrative generation without retraining. During \textit{answer inference}, the image, generated narrative, and task-specific question prompt are utilized to derive answers for different tasks. As gender information is obfuscated in generated narratives, the model is encouraged to rethink gender attributes in this stage, which contributes to more appropriate and unbiased answers.

We conduct extensive experiments to demonstrate the effectiveness of GAMA. First, GAMA is evaluated across two vision-language tasks: image captioning and image search. Additionally, zero-shot experiments are performed on two benchmarks for measuring gender bias, namely VisoGender \citep{hall_visogender_2023} and VL-Bias \citep{zhang_counterfactually_2022}. Experimental results show that GAMA performs well against previous debiasing methods on task performance and gender bias mitigation. Notably, GAMA exhibits remarkable generalization ability on both benchmarks. Moreover, we showcase the effectiveness of our proposed modules in reducing object hallucination and gender bias. For further explanation, we probe the connection between object hallucination and gender bias. 

\section{Related Work}

\begin{figure*}[t]
    \centering
    \includegraphics[width=\linewidth]{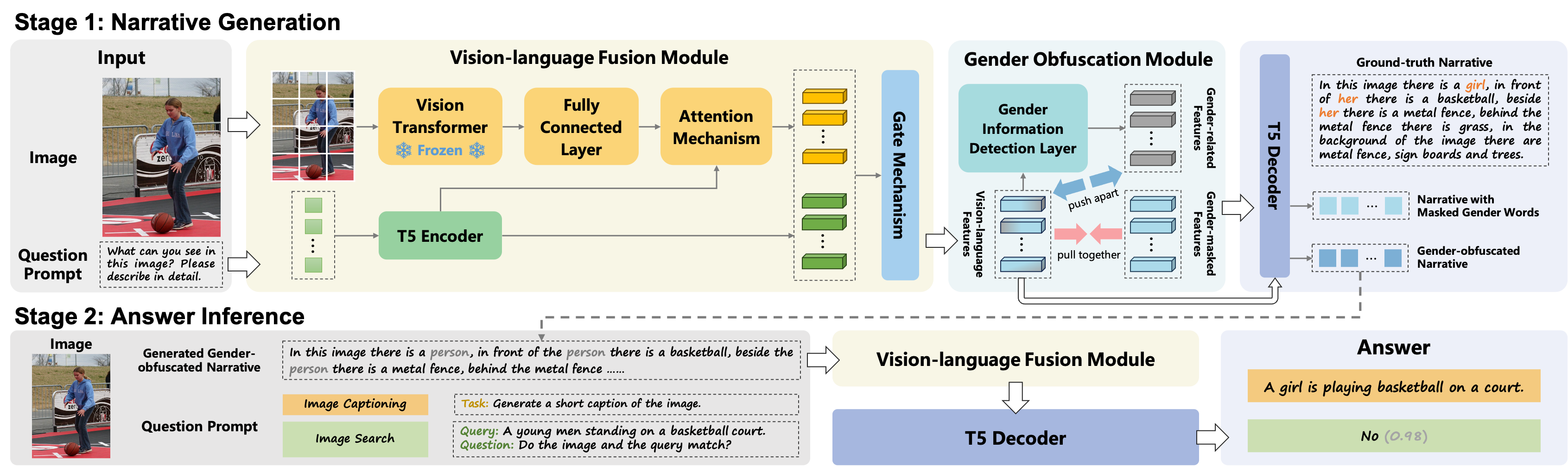}
    \caption{\textbf{The overall framework of GAMA.} We briefly provide task-specific question prompts and answers, which are detailed in Appendix \ref{app:impl-detail}. We take the token probability of the decoder as the match score in image search.}
    \label{fig:model}
\end{figure*}

\subsection{Sources of gender bias}

The growing popularity of multi-modality has prompted research into sources of gender bias in vision-language tasks. First, \textit{datasets} are a prominent factor, as object labeling is closely linked to our conceptualization \citep{brown_how_1958}. Thus real-life stereotypes subtly infiltrate datasets. \citet{hirota_gender_2022} and \citet{harrison_run_2023} highlighted the underrepresentation of women in datasets. Second, \textit{pre-trained VLMs} (PVLMs) have enhanced vision-language tasks by leveraging extensive knowledge from pre-training data. Consequently, PVLMs inherit such bias from both language and vision sources \citep{wang_balanced_2019,ross_measuring_2021,srinivasan_worst_2022}. Third, \textit{model structures} may amplify gender bias in datasets \citep{zhao_men_2017}. Researchers have observed stereotype exaggeration in various task-specific VLMs \citep{kay_unequal_2015,bhargava_exposing_2019,wang_are_2021}.

Given these diverse bias sources, gender bias may not be effectively addressed by merely pre-processing data or debiasing pre-training features.

\subsection{Gender bias mitigation}

Research interest in gender bias mitigation for vision-language tasks has notably increased. Previous studies addressed gender bias in unimodal models, such as language \citep{he_mabel_2022,shaikh_second_2023} or vision \citep{wang_balanced_2019,steed_image_2021}. 
\citet{zhou_vlstereoset_2022} uncovered both intra-modal and inter-modal gender bias. Previous work on gender bias mitigation for VLMs can be divided into three classes: 1) \textit{adding debiasing modules to existing task-specific VLMs}, which aims to pre-process imbalanced training sets \citep{zhao_men_2017}, process intermediate features \citep{hendricks_women_2018,wang_are_2021}, or post-process biased model outputs \citep{hirota_model-agnostic_2023,janghorbani_multi-modal_2023}; 2) \textit{re-designing training objectives of PVLMs} to produce debiased features through negative \citep{wang_balanced_2019} or counterfactual \citep{zhang_counterfactually_2022} samples; and 3) \textit{proposing new model structures} designed to learn fair representations by capturing gender visual evidence \citep{tang_mitigating_2021}, or adding a negative residual \citep{seth_dear_2023}. However, previous studies superficially address gender bias by removing biased data or features and ignore the correlation between gender bias and object hallucination. 

\subsection{Object hallucination}

Despite the great success of VLMs in vision-language tasks, they still suffer from object hallucination.
Object hallucination refers to the fact that the content generated by the model contains objects inconsistent with or absent from the given image \citep{rohrbach_object_2018}.
\citet{rohrbach_object_2018} proposed two metrics, CHAIRs and CHAIRi, to measure object hallucination. A recent approach, POPE \citep{li_evaluating_2023}, is a polling-based query method with enhanced flexibility.
Efforts to mitigate object hallucination in VLMs fall into two categories. Some methods aim to disentangle object co-occurrence patterns. For instance, \citet{biten_let_2022} introduced object labels and altered objects in the captions, while \citet{zhou_analyzing_2023} replaced error-prone objects with placeholder tags. Others focus on minimizing logical errors, such as leveraging chain-of-thought (CoT) prompting to generate intermediate reasoning chains as rationales \citep{zhang_multimodal_2023}.

\section{Method}

In this section, we will ﬁrst overview the framework of GAMA and then introduce the training data. Finally, we detail our model architecture. 

\subsection{Overview}

GAMA is composed of two stages: narrative generation and answer inference, as illustrated in Figure \ref{fig:model}. During narrative generation, GAMA takes an image and a question prompt as input to yield an all-sided but gender-obfuscated narrative. During answer inference, different vision-language tasks are formulated as generation tasks. GAMA utilizes the image, generated narrative, and task-specific question prompt to generate a task-specific answer. We formulate both stages as taking an image $I$ and a text sequence $X=[x_1, x_2, \cdots, x_n]$ with $n$ tokens as input and outputting a target sequence $Y=[y_1, y_2, \cdots, y_m]$ with $m$ tokens. 

The models in the two stages are trained independently. The trained narrative generation model can be applied to different tasks without retraining. 

GAMA comprises three key modules: a vision-language fusion module to extract features in both stages, a gender obfuscation module to mask gender-related information during narrative generation, and a decoder for answer generation.

\subsection{Training data}

During narrative generation, GAMA strives to generate an all-sided narrative for the given image to avoid premature focus on localized details. 
To this end, we train the model with the Localized Narratives dataset \citep{voigtlaender_connecting_2023} for Open Images \citep{kuznetsova_open_2020}. The dataset is crafted to depict image regions covered by a mouse trace, associating specific image regions with specific words in the narrative. Consequently, the narratives are spread throughout the entire image, rather than confined to specific localized regions. 

\subsection{Vision-language fusion module}

The vision-language fusion module is designed to learn vision-language representations in two stages.

\paragraph{Encoder}

We use the T5 \citep{raffel_exploring_2020} encoder to learn language representations, and a Vision Transformer (ViT) \citep{dosovitskiy_image_2020} to extract vision features:
\begin{gather} \label{a}
    \mathbf{H}_l = \textrm{T5Encoder}([x_1, x_2, \cdots, x_n]), \\
    \mathbf{H}_v = \mathbf{W}_1 \textrm{ViT} (I),
\end{gather}
where $\mathbf{H}_l, \mathbf{H}_v \in \mathbb{R}^{n\times d}$, and $d$ is the hidden dimension. $\mathbf{W}_1$ is a projection matrix to convert the shape of features extracted by the ViT. Notably, the parameters of the ViT are frozen.

\paragraph{Feature fusion}
We utilize an attention mechanism \citep{vaswani_attention_2017} to capture the interactions between language and vision features:
\begin{equation}
    \mathbf{\hat{H}}_v = \textrm{Attention}(\mathbf{H}_l, \mathbf{H}_v, \mathbf{H}_v).
\end{equation}

We aggregate $\mathbf{H}_l$ and $\mathbf{\hat{H}}_v$ with a gate mechanism \citep{li_vision_2022, fang_neural_2022}. The gate $\lambda$ and the fused features $\mathbf{H}$ are calculated as:
\begin{gather}
    \lambda = \textrm{Sigmoid} (\mathbf{W}_2 \mathbf{H}_l + \mathbf{W}_3 \mathbf{\hat{H}}_v), \\
    \mathbf{H} = (1-\lambda) \cdot  \mathbf{H}_l + \lambda \cdot \mathbf{\hat{H}}_v,
\end{gather}
where $\mathbf{W}_2$ and $\mathbf{W}_3$ are trainable parameters.

\subsection{Gender obfuscation module} \label{subsec:gender-obfucation-module}

The gender obfuscation module is employed during narrative generation to obfuscate gender information in narratives. This module prevents the model from biasing the context due to gender attributes. 

\paragraph{Pre-processing}
First, we replace gender words with a special token \texttt{[GENDER]} and leave other words unchanged to obtain a narrative with masked gender words $\bar{Y}=[\bar{y}_1, \bar{y}_2, \cdots, \bar{y}_m]$.\footnote{The gender word list comes from \citet{hirota_model-agnostic_2023}, which can be found in Appendix \ref{app:gender-word}.}

\paragraph{Gender information detection}

Then we construct the gender-masked features $\mathbf{\bar{H}}$. We detect gender-related features $\mathbf{H}_g$ and remove them from the vision-language features $\mathbf{H}$ as follows:
\begin{gather}
    \mu = \textrm{Sigmoid} (\mathbf{W}_4 \mathbf{H}), \\
    \mathbf{H}_g = \mu \cdot \mathbf{H}, \\
    \mathbf{\bar{H}} = \mathbf{H} - \mathbf{H}_g,
\end{gather}
where $\textbf{W}_4$ is a trainable parameter.

\paragraph{Contrastive loss}
Motivated by contrastive learning \citep{he_momentum_2020, chen_simple_2020}, we employ a contrastive loss to keep the vision-language features $\mathbf{H}$ close to the gender-masked features $\mathbf{\bar{H}}$ and away from the gender-related features $\mathbf{H}_g$. This approach helps obscure gender details as well as preserve context information. 
We define our contrastive loss as:
\begin{equation}
    \mathcal{L}_{con} = - \log 
    \frac{e^{\textrm{s}(\mathbf{H},\mathbf{\bar{H}}) / \tau}}
    {e^{\textrm{s}(\mathbf{H},\mathbf{\bar{H}}) / \tau} + 
    e^{\textrm{s}(\mathbf{H},\mathbf{H}_g) / \tau}},
\end{equation}
where $\textrm{s}(\mathbf{U}, \mathbf{V}) = \mathbf{U}^{\top}\mathbf{V} / \|\mathbf{U}\| \|\mathbf{V}\|$ denotes the cosine similarity between $\mathbf{U}$ and $\mathbf{V}$, and $\tau$ is a temperature hyper-parameter.

\subsection{Decoder}

The T5 decoder takes the encoder output $\mathbf{\tilde{H}}$, text sequence $X$, and previously generated tokens $\tilde{Y}_{<t}$ as inputs to get the $t$-th token probability distribution:
\begin{gather}
    \mathbf{\tilde{h}}_t = \textrm{T5Decoder}(\mathbf{\tilde{H}},X,\tilde{Y}_{<t}), \\
    P(\tilde{y}_t | \mathbf{\tilde{H}},X,\tilde{Y}_{<t}) = \textrm{Softmax} (\mathbf{W}_5 \mathbf{\tilde{h}}_t),
\end{gather}
where $\mathbf{W}_5$ is a trainable parameter to map the hidden dimension to the vocabulary size. For each symbol $\tilde{a} \in \{\mathbf{\tilde{H}}, \mathbf{\tilde{h}}_t, \tilde{Y}_{<t}, \tilde{y}_{t} \}$, $\tilde{a}$ is either $a$ or $\bar{a}$.

\subsection{Training objective}

During the training phase, we use the teacher forcing to train the models in the two stages. The models in two stages are trained independently.

\paragraph{Narrative generation}

In the narrative generation stage, we train the model with the cross-entropy loss on ground-truth narratives and narratives with masked gender words, which can be defined as:
\begin{gather}
    \mathcal{L}_{ce} = - \sum_{t=1}^{m}\log {P(y_t | \mathbf{H},X,Y_{<t})}, \\
    \mathcal{\bar{L}}_{ce} = - \sum_{t=1}^{m}\log {P(\bar{y}_t | \mathbf{\bar{H}},X,\bar{Y}_{<t})}.
\end{gather}

In total, the narrative generation stage contains three losses, which can be formulated as:
\begin{equation}
    \mathcal{L}_1 = \mathcal{L}_{con} + \mathcal{L}_{ce} + \mathcal{\bar{L}}_{ce}.
\end{equation}

\paragraph{Answer inference}

The answer inference model is trained with a cross-entropy loss:
\begin{equation}
    \mathcal{L}_2 = - \sum_{t=1}^{m}\log {P(y_t | \mathbf{H},X,Y_{<t})}.
\end{equation}

\section{Experiment Setup}

To demonstrate the effectiveness and generalization ability of GAMA, we conduct extensive experiments across two vision-language tasks (image captioning and image search) and two benchmarks for measuring gender bias (VisoGender \citep{hall_visogender_2023} and VL-Bias \citep{zhang_counterfactually_2022}).\footnote{Due to space constraints, experiments on VL-Bias can be found in Appendix \ref{app:vqa}.}

\subsection{Datasets} \label{subsec:exp-dataset}

The statistics of the datasets are listed in Table \ref{tab:data-stat}.

\paragraph{Gender label}

Following previous work \citep{hendricks_women_2018,wang_are_2021,hirota_model-agnostic_2023}, we utilize ground-truth captions to label the gender attributes of images. Specifically, an image will be labeled as ``male (female)'' if at least one of its captions contains male (female) gender words and no captions contain female (male) words. Otherwise, it will be labeled as ``neutral''.

\paragraph{Narrative generation}

The Localized Narratives dataset \citep{voigtlaender_connecting_2023} for Open Images \citep{kuznetsova_open_2020} is utilized to train the narrative generation model.\footnote{We ascertain the absence of any overlap between Localized Narratives and other test sets through the cosine similarity among image tensors with a threshold of $0.9$. See Appendix \ref{app:ablation} for further studies on the training set size of Localized Narratives.}

\paragraph{Image captioning}

We experiment on MSCOCO captions \citep{chen_microsoft_2015}.
Following \citet{hirota_model-agnostic_2023}, we use the original MSCOCO training set for training, a subset of the MSCOCO validation set from \citet{zhao_understanding_2021} for test,\footnote{Access the test split at \url{https://princetonvisualai.github.io/imagecaptioning-bias/}.} and the remaining images for validation. Each image is associated with five human-annotated captions. Notably, no images in the test set are labeled as ``neutral''.

\paragraph{Image search} 

Our evaluation for image search involves the MSCOCO \citep{chen_microsoft_2015} and Flickr30K \citep{young_image_2014} datasets.
Following \citet{wang_are_2021}, we use the Karpathy split \citep{karpathy_deep_2015} for training and validation. As test sets in the Karpathy split are gender-unbalanced, we randomly select the same number of images with their gender-neutral queries \footnote{The queries are available at \url{https://github.com/eric-ai-lab/Mitigate-Gender-Bias-in-Image-Search}.} under ``male'', ``female'' and ``neutral'' labels for test.

\begin{table}
\centering
\resizebox{\columnwidth}{!}{%
\begin{tabular}{cccccc} 
\toprule
\textbf{Stage}     & \textbf{Task}                                                           & \textbf{Dataset}                                               & \textbf{Train} & \textbf{Dev} & \textbf{Test}  \\ 
\midrule
1                  & \begin{tabular}[c]{@{}c@{}}Narrative\\ Generation\end{tabular}          & \begin{tabular}[c]{@{}c@{}}Localized\\ Narratives\end{tabular} & 507,444        & 41,691       & 126,020        \\ 
\midrule
\multirow{4}{*}{2} & \begin{tabular}[c]{@{}c@{}}Image\\ Captioning\end{tabular}              & MSCOCO                                                         & 82,783         & 29,724       & 10,780         \\ 
\cmidrule(lr){2-6}
                   & \multirow{2}{*}{\begin{tabular}[c]{@{}c@{}}Image\\ Search\end{tabular}} & MSCOCO                                                         & 113,287        & 5,000        & 1,500          \\
                   &                                                                         & Flickr30K                                                      & 29,000         & 1,000        & 300            \\ 
\cmidrule(lr){2-6}
                   & \multicolumn{2}{c}{VisoGender}                                                                                                           & /              & /            & 690            \\
\bottomrule
\end{tabular}
}
\caption{\textbf{The statistics of datasets.} We show the counts of images within each split of the datasets.}
\label{tab:data-stat}
\end{table}

\paragraph{VisoGender}

VisoGender \citep{hall_visogender_2023} benchmarks occupation-related gender bias in VLMs. VLMs are required to align images with correct gender pronouns in a resolution task and retrieve the top-K images for a gender-neutral caption of a given occupation in a retrieval task.

\subsection{Baselines}

\paragraph{Image captioning}

GAMA is compared with the following methods for image captioning: 1) \textbf{Equalizer} \citep{hendricks_women_2018}, which focuses on the ``person'' segmentation to make gender-speciﬁc predictions; 2) \textbf{GAIC$_{es}$} \citep{tang_mitigating_2021}, which encourages capturing gender visual evidence through self-guided visual attention; and 3) \textbf{LIBRA} \citep{hirota_model-agnostic_2023}, which leverages state-of-the-art captioning models to generate high-quality captions and debiases through an additional editing model.

\paragraph{Image search} 

We evaluate GAMA against the following methods for image search: 1) \textbf{SCAN-FS} \citep{wang_are_2021}, which applies a fair sampling method to the representative image search baseline SCAN \citep{lee_stacked_2018}; 2) \textbf{CLIP-clip} \citep{wang_are_2021}, which introduces a feature pruning algorithm to the features generated by the PVLM CLIP \citep{radford_learning_2021}; and 3) \textbf{FairVLP} \citep{zhang_counterfactually_2022}, which trains the PVLM ALBEF \citep{li_align_2021} with counterfactual samples to obtain debiased representations.

\paragraph{VisoGender}

Following \citet{hall_visogender_2023}, we evaluate GAMA against state-of-the-art pre-trained vision-language encoders (\textbf{CLIP} \citep{radford_learning_2021}, \textbf{OpenCLIP} \citep{cherti_reproducible_2023} trained on LAION 2B and 400M, \textbf{SLIP} \citep{mu_slip_2022}, \textbf{DeCLIP} \citep{li_supervision_2022} and \textbf{FILIP} \citep{yao_filip_2022}) and pre-trained captioning models (\textbf{BLIP-2} \citep{li_blip-2_2023} and \textbf{GIT} \citep{wang_git_2022}).

\begin{table*}[h]
\centering
\resizebox{0.8\textwidth}{!}{%
\begin{tabular}{cccccccccc} 
\toprule
\multicolumn{2}{c}{\multirow{2}{*}{\textbf{Model}}}                                              & \multicolumn{3}{c}{\textbf{Gender Bias Metrics $\downarrow$}}                                                                  & \multicolumn{5}{c}{\textbf{Image Captioning Metrics $\uparrow$}}                                                                                                                                                       \\ 
\cmidrule(lr){3-5}\cmidrule(lr){6-10}
\multicolumn{2}{c}{}                                                                             & \textbf{LIC}                             & \textbf{Error}                          & \textbf{BiasAmp}                          & \textbf{BLEU-4}                          & \textbf{CIDEr}                            & \textbf{METEOR}                          & \textbf{SPICE}                           & \textbf{CLIPScore}                        \\ 
\midrule
\multicolumn{2}{c}{Equalizer $^\dag$}                                                            & 0.7                                      & 8.1                                     & -0.50                                     & 27.2                                     & 79.8                                      & 24.1                                     & 16.8                                     & 69.9                                      \\
\multicolumn{2}{c}{GAIC$_{es}$ $^\dag$}                                                          & 1.4                                      & 5.9                                     & -0.70                                     & 32.6                                     & 94.5                                      & 27.4                                     & 18.3                                     & 72.7                                      \\ 
\midrule
NIC \citep{vinyals_show_2015}              & \multirow{9}{*}{+LIBRA $^\flat$} & -0.3                                     & 5.7                                     & -1.47                                     & 24.6                                     & 72.0                                      & 24.2                                     & 16.5                                     & 71.7                                      \\
SAT \citep{xu_show_2015}                   &                                  & -1.4                                     & {\cellcolor[rgb]{0.851,0.882,0.949}}3.9                                     & -0.48                                     & 34.6                                     & 95.9                                      & 27.8                                     & 20.0                                     & 73.6                                      \\
FC \citep{rennie_self-critical_2017}       &                                  & -0.2                                     & 4.3                                     & -1.11                                     & 32.8                                     & 95.9                                      & 27.3                                     & 19.7                                     & 72.9                                      \\
Att2in \citep{rennie_self-critical_2017}   &                                  & -0.3                                     & 4.6                                     & {\cellcolor[rgb]{0.851,0.882,0.949}}-3.39 & 35.9                                     & 101.7                                     & 28.5                                     & 20.6                                     & 73.8                                      \\
UpDn \citep{anderson_bottom-up_2018}       &                                  & 1.5                                      & 4.5                                     & -2.23                                     & 37.7                                     & 110.1                                     & 29.6                                     & 22.0                                     & 74.6                                      \\
Transformer \citep{vaswani_attention_2017} &                                  & 2.3                                      & 5.0                                     & -0.26                                     & 33.9                                     & 98.7                                      & 28.6                                     & 20.9                                     & 75.7                                      \\
OSCAR \citep{li_oscar_2020}                &                                  & 0.3                                      & 4.6                                     & -1.95                                     & 37.2                                     & 113.1                                     & {\cellcolor[rgb]{0.886,0.937,0.855}}31.1 & {\cellcolor[rgb]{0.886,0.937,0.855}}23.2 & 75.7                                      \\
ClipCap \citep{mokady_clipcap_2021}        &                                  & {\cellcolor[rgb]{0.886,0.937,0.855}}-1.5 & 4.5                                     & -0.57                                     & 33.8                                     & 100.6                                     & 29.3                                     & 21.4                                     & {\cellcolor[rgb]{0.886,0.937,0.855}}76.0  \\
GRIT \citep{nguyen_grit_2022}              &                                  & 0.7                                      & 4.1 & 1.57                                      & {\cellcolor[rgb]{0.886,0.937,0.855}}40.5 & {\cellcolor[rgb]{0.886,0.937,0.855}}116.8 & 30.6                                     & 22.6 & {\cellcolor[rgb]{0.851,0.882,0.949}}75.9  \\ 
\midrule
\multicolumn{2}{c}{\textbf{GAMA}}                                                                & {\cellcolor[rgb]{0.851,0.882,0.949}}-1.1 & {\cellcolor[rgb]{0.886,0.937,0.855}}3.4 & {\cellcolor[rgb]{0.886,0.937,0.855}}-3.40 & {\cellcolor[rgb]{0.851,0.882,0.949}}38.2 & {\cellcolor[rgb]{0.851,0.882,0.949}}115.1 & {\cellcolor[rgb]{0.851,0.882,0.949}}31.0 & {\cellcolor[rgb]{0.851,0.882,0.949}}22.7                                     & 75.4                                      \\
\bottomrule
\end{tabular}
}
\caption{\textbf{Results of image captioning.} The best results are highlighted in \sethlcolor{green-table}\hl{green}, and the second-best are in \sethlcolor{blue-table}\hl{blue}. For gender bias metrics, lower is better. For image captioning metrics, higher is better. Gender bias metrics are scaled by $100$. $\dag$: the results are reproduced with official implementation; $\flat$: the results are retrieved from \citet{hirota_model-agnostic_2023}.}

\label{tab:caption-result}
\end{table*}

\begin{table}[h]
\centering
\resizebox{\columnwidth}{!}{%
\begin{tabular}{cccccccc} 
\toprule
\multirow{2}{*}{\textbf{Dataset}}   & \multirow{2}{*}{\textbf{Model}} & \multicolumn{3}{c}{\textbf{Gender Bias Metrics}}                                                                                     & \multicolumn{3}{c}{\textbf{Image Search Metrics $\uparrow$}}                                                                    \\ 
\cmidrule(l){3-8}
                                    &                                 & \textbf{Bias@1}                            & \textbf{Bias@5}                            & \textbf{Bias@10}                           & \textbf{Recall@1}                        & \textbf{Recall@5}                        & \textbf{Recall@10}                        \\ 
\midrule
\multirow{4}{*}{\textbf{MSCOCO}}    & SCAN-FS                         & -0.1043                                    & -0.1716                                    & -0.2392                                    & 25.4                                     & 54.8                                     & 66.1                                      \\
                                    & CLIP-clip                       & -0.1173                                    & -0.1940                                    & -0.2528                                    & 28.9                                     & 57.2                                     & 68.5                                      \\
                                    & FairVLP                         & {\cellcolor[rgb]{0.851,0.882,0.949}}0.0334 & {\cellcolor[rgb]{0.851,0.882,0.949}}0.1293 & {\cellcolor[rgb]{0.886,0.937,0.855}}0.1965 & {\cellcolor[rgb]{0.851,0.882,0.949}}58.7 & {\cellcolor[rgb]{0.851,0.882,0.949}}80.2 & {\cellcolor[rgb]{0.851,0.882,0.949}}90.4  \\
                                    & \textbf{GAMA}                   & {\cellcolor[rgb]{0.886,0.937,0.855}}0.0273 & {\cellcolor[rgb]{0.886,0.937,0.855}}0.1281 & {\cellcolor[rgb]{0.851,0.882,0.949}}0.1995 & {\cellcolor[rgb]{0.886,0.937,0.855}}63.7 & {\cellcolor[rgb]{0.886,0.937,0.855}}83.6 & {\cellcolor[rgb]{0.886,0.937,0.855}}93.5  \\ 
\midrule
\multirow{4}{*}{\textbf{Flickr30K}} & SCAN-FS                         & -0.1281                                    & -0.1857                                    & -0.2469                                    & 36.4                                     & 67.6                                     & 78.3                                      \\
                                    & CLIP-clip                       & -0.1050                                    & -0.1603                                    & -0.2307                                    & 64.0                                     & 86.5                                     & 91.9                                      \\
                                    & FairVLP                         & {\cellcolor[rgb]{0.851,0.882,0.949}}0.0514 & {\cellcolor[rgb]{0.851,0.882,0.949}}0.1012 & {\cellcolor[rgb]{0.851,0.882,0.949}}0.1731 & {\cellcolor[rgb]{0.851,0.882,0.949}}77.4 & {\cellcolor[rgb]{0.851,0.882,0.949}}95.2 & {\cellcolor[rgb]{0.851,0.882,0.949}}97.1  \\
                                    & \textbf{GAMA}                   & {\cellcolor[rgb]{0.886,0.937,0.855}}0.0449 & {\cellcolor[rgb]{0.886,0.937,0.855}}0.0941 & {\cellcolor[rgb]{0.886,0.937,0.855}}0.1676 & {\cellcolor[rgb]{0.886,0.937,0.855}}83.1 & {\cellcolor[rgb]{0.886,0.937,0.855}}95.8 & {\cellcolor[rgb]{0.886,0.937,0.855}}97.9  \\
\bottomrule
\end{tabular}
}
\caption{\textbf{Results of image search.} For gender bias metrics, closer to $0$ is better. Baselines are reproduced with official implementation. We report the average across $3$ runs. Note that we evaluate models with gender-balanced test sets instead of the Karpathy test sets utilized by \citet{wang_are_2021} and \citet{zhang_counterfactually_2022}.}
\label{tab:search-result}
\end{table}

\subsection{Metrics} 

We present metrics for evaluation, covering task performance, gender bias, and object hallucination. Calculation details are presented in Appendix \ref{app:metrics}.

\textbf{Task performance metrics} are used to evaluate model performance on specific tasks.
For \textit{image captioning}, we use established referenced-based metrics BLEU-4 \citep{papineni_bleu_2002}, CIDEr \citep{vedantam_cider_2015}, METEOR \citep{denkowski_meteor_2014} and SPICE \citep{anderson_spice_2016} as well as a reference-free metric CLIPScore \citep{hessel_clipscore_2021}.
For \textit{image search}, we employ Recall@K to calculate the ratio of correct images in the top-K retrieved images.

\textbf{Gender bias metrics} are employed to measure the model performance on gender bias mitigation.
For \textit{image captioning}, we use LIC \citep{hirota_quantifying_2022} to measure gender bias in the context of gender words, Error \citep{hendricks_women_2018} to assess the gender misclassification ratio of generated captions, and BiasAmp \citep{zhao_men_2017} to quantify bias amplification based on word-gender co-occurrence.
For \textit{image search}, we follow \citet{wang_are_2021} and adopt Bias@K to measure gender bias among the top-K images. 
For \textit{VisoGender}, we follow the setting in \citet{hall_visogender_2023}. The resolution accuracy gap ($\Delta_{RA}$) is used to measure resolution bias, while Bias@K \citep{wang_are_2021}, Skew@K \citep{geyik_fairness-aware_2019} and NDKL \citep{geyik_fairness-aware_2019} are used to measure retrieval bias.

\textbf{Object hallucination metrics} are utilized to probe the connection between object hallucination and gender bias.
CHAIRs and CHAIRi \citep{rohrbach_object_2018} are used to evaluate incorrect object generation at the sentence and object levels, respectively. \citet{rohrbach_object_2018} used a synonym list \citep{lu_neural_2018} to map words to MSCOCO objects. We refine it with hierarchical object relationships.\footnote{Details can be found in Appendix \ref{app:hallucination-metrics}.}

\subsection{Implementation details}

We use \texttt{flan-t5-base} \citep{chung_scaling_2022} as the backbone. Our image encoder is \texttt{vit-base- patch16-384} \citep{dosovitskiy_image_2020}, and its parameters are frozen during training. For narrative generation, we set the temperature $\tau=0.1$ and conduct further studies in Appendix \ref{app:ablation}. Experiments on VisoGender are conducted on GAMA search models under a zero-shot setting. More details are presented in Appendix \ref{app:impl-detail}.

\section{Results and Analysis}


In this section, we seek to answer the following research questions: \textbf{RQ1:} Does GAMA perform well across different vision-language tasks in task performance metrics and gender bias metrics? (Section \ref{subsec:overall-result}) \textbf{RQ2:} Do our gender obfuscation module and two-stage framework help GAMA think before acting? (Section \ref{subsec:ablation-study}) \textbf{RQ3:} Are object hallucination and gender bias closely related? (Section \ref{subsec:ablation-study}) \textbf{RQ4:} Can GAMA be generalized to specialized datasets for gender bias measurement under a zero-shot setting? (Section \ref{subsec:generalization})

\subsection{Overall results} \label{subsec:overall-result}

We conduct experiments on GAMA and baselines. Results on image captioning and image search are shown in Table \ref{tab:caption-result} and Table \ref{tab:search-result}, respectively. 

To begin with, GAMA outperforms or closely aligns with the baselines in both two tasks. It is noteworthy that most of these baselines are tailored to a specific task. In contrast, GAMA is a task-agnostic method and can be easily applied to different vision-language tasks. 

Then we compare GAMA with each baseline separately. First, GAMA comfortably outperforms Equalizer and GAIC$_{es}$. As Equalizer forces the model to focus on persons in images, it loses necessary information to generate the correct context of gender words. Similarly, GAIC$_{es}$ is designed to improve gender classification accuracy with additional gender evide nce. Therefore, these two baselines prove less effective.

Second, LIBRA achieves noticeable results with different captioning models. LIBRA considers gender bias manifested as gender misclassification as well as biased context. However, there are some problems. 1) LIBRA is limited by the performance of captioning models, as it is designed to revise their outputs. 2) LIBRA is designed to edit biased captions after a captioning model, but it is trained with synthesized data instead of actual model outputs, potentially causing error propagation. Consequently, the performance of LIBRA depends on the specific captioning model in use, rendering its results on gender bias metrics less stable.

Third, GAMA also performs better than SCAN-FS, CLIP-clip and FairVLP in image search. We observe that SCAN-FS and CLIP-clip exhibit overcorrection, leading to the underrepresentation of males. These baselines tackle gender bias mainly by removing biased data or features. Given the diverse sources of bias, it cannot be effectively mitigated by merely processing data or debiasing pre-training features.

\begin{figure}
    \centering
    \includegraphics[width=0.6\linewidth]{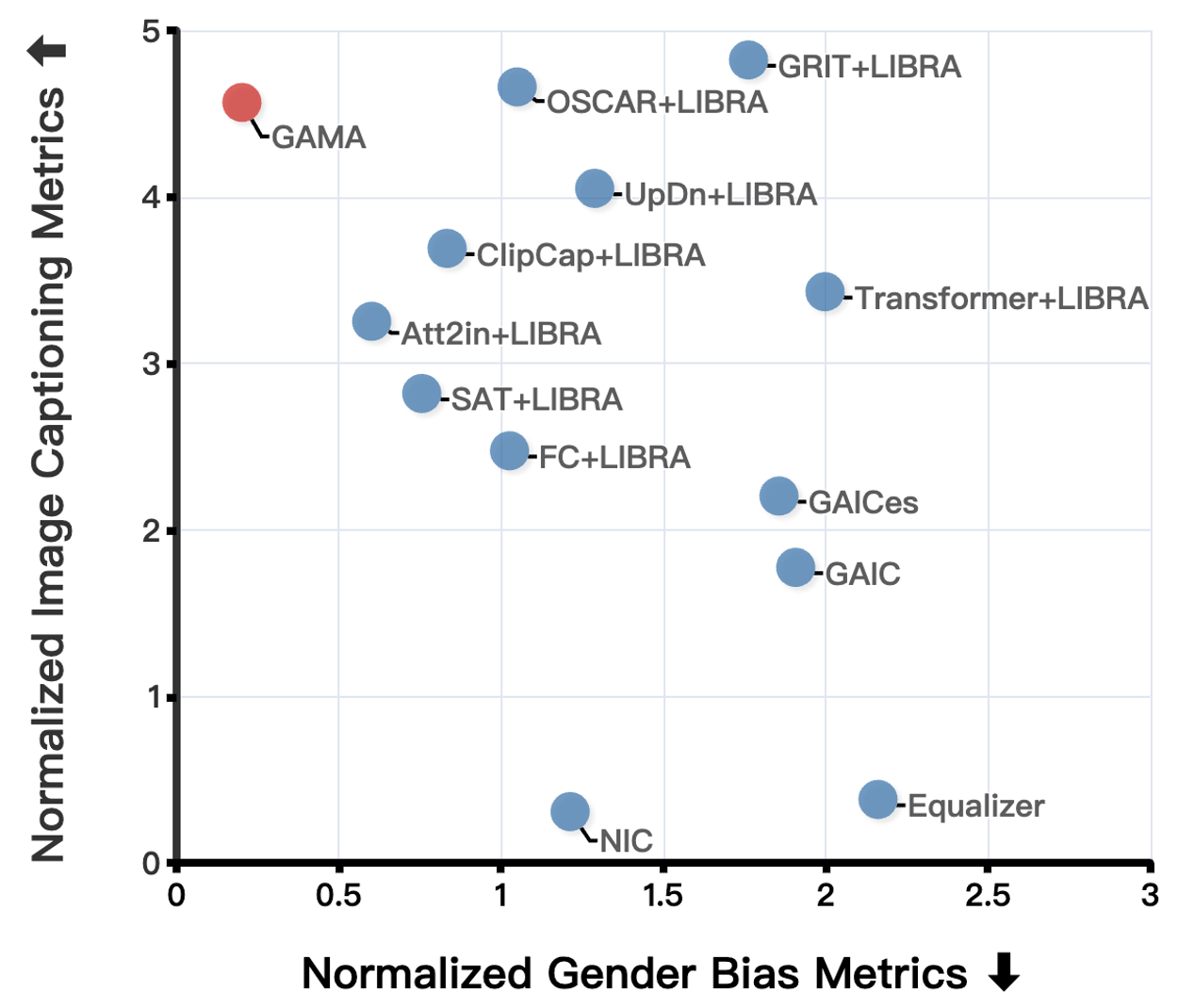}
    \caption{\textbf{Comparison on task performance and gender bias mitigation ability.} We normalize the metrics separately and sum the normalized gender bias metrics and image captioning metrics, respectively.}
    \label{fig:exp-trade-off}
\end{figure}

Finally, we compare models on task performance and gender bias mitigation ability in Figure \ref{fig:exp-trade-off}. Results show that GAMA strikes a good balance between task performance and gender bias mitigation. 

\begin{table*}
\centering
\resizebox{0.9\linewidth}{!}{%
\begin{tabular}{ccccccccc} 
\toprule
\multirow{2}{*}{\textbf{Model}} & \multicolumn{3}{c}{\textbf{Gender Bias Metrics $\downarrow$}} & \textbf{Gender-related Statistics $\downarrow$} & \multicolumn{2}{c}{\textbf{Object Hallucination Metrics $\downarrow$ }} & \multicolumn{2}{c}{\textbf{Co-occurrence Statistics $\downarrow$}}  \\ 
\cmidrule(lr){2-4}\cmidrule(lr){5-5}\cmidrule(lr){6-7}\cmidrule(lr){8-9}
                                & \textbf{LIC} & \textbf{Error} & \textbf{BiasAmp}              & \textbf{\#Gender}                  & \textbf{CHAIRs} & \textbf{CHAIRi}                                       & \textbf{HR$_C^g$@10} & \textbf{HR$^o_C$@10}            \\ 
\midrule
\textbf{GAMA}                   & -1.1         & 3.4            & -3.40                         & 55.61                              & 10.94           & 6.02                                                  & 38.15                & 50.30                           \\ 
\midrule
w/o GO $^\blacklozenge$         & 0.5          & 3.1            & -1.26                         & 62.49                              & 12.81           & 7.36                                                  & 41.52                & 50.09                           \\
w/o narrative $^\bigstar$       & -0.6         & 4.1            & -2.87                         & /                                  & 13.02           & 7.84                                                  & 41.37                & 51.94                           \\
single-stage~$^\clubsuit$       & -0.9         & 3.6            & -2.13                         & /                                  & 12.40           & 7.02                                                  & 39.91                & 51.76                           \\
\bottomrule
\end{tabular}
}
\caption{\textbf{Results of ablation studies.} \#Gender denotes the proportion of generated narratives with gender words in the test set. Note that no images are labeled as ``neutral'' in the test set.
CHAIRs and CHAIRi are calculated based on our refined synonym list. HR$_C^o$@10 \citep{li_evaluating_2023} and HR$_C^g$@10 are used to quantify whether the model is prone to hallucinate objects that frequently co-occur with ground-truth objects and gender attributes, respectively, which are detailed in Appendix \ref{app:HR}.
$^\blacklozenge$: the gender obfuscation module (GO) is removed from narrative generation;
$^\bigstar$: the model is trained to generate gender-obfuscated captions instead of gender-obfuscated narratives in the first stage;
$^\clubsuit$: the narrative generation stage is removed, and GO is adopted in the answer inference stage.
Note that the single-stage framework does not incorporate narratives as input.
}
\label{tab:ablation-study}
\end{table*}

\begin{table*}[h]
\centering
\resizebox{0.8\textwidth}{!}{%
\begin{tabular}{ccccccccccccccc} 
\toprule
\multicolumn{2}{c}{\multirow{3}{*}{\textbf{Model}}} & \multicolumn{3}{c}{\textbf{Resolution}}                                                                                              & \multicolumn{10}{c}{\textbf{Retrieval}}                                                                                                                                                                                                                                                                                                                                                                                                       \\ 
\cmidrule(lr){3-5}\cmidrule(lr){6-15}
\multicolumn{2}{c}{}                                & \multirow{2}{*}{\textbf{Accuracy~$\uparrow$}} & \multicolumn{2}{c}{\textbf{$\Delta_{RA}$}}                                           & \multicolumn{2}{c}{\textbf{Bias@5}}                                                 & \multicolumn{2}{c}{\textbf{Bias@10}}                                                & \multicolumn{2}{c}{\textbf{MaxSkew@5~$\downarrow$}}                                 & \multicolumn{2}{c}{\textbf{MaxSkew@10~\textbf{$\downarrow$}}}                       & \multicolumn{2}{c}{\textbf{NDKL~\textbf{$\downarrow$}}}                              \\ 
\cmidrule(lr){4-5}\cmidrule(lr){6-7}\cmidrule(lr){8-9}\cmidrule(lr){10-11}\cmidrule(lr){12-13}\cmidrule(lr){14-15}
\multicolumn{2}{c}{}                                &                                               & \textbf{OO}                              & \textbf{OP}                               & \textbf{Mean}                            & \textbf{$\sigma$}                        & \textbf{Mean}                            & \textbf{$\sigma$}                        & \textbf{Mean}                            & \textbf{$\sigma$}                        & \textbf{Mean}                            & \textbf{$\sigma$}                        & \textbf{Mean}                            & \textbf{$\sigma$}                         \\ 
\midrule
\multicolumn{2}{c}{CLIP}                            & 0.75                                          & -0.14                                    & -0.27                                     & 0.11                                     & {\cellcolor[rgb]{0.886,0.937,0.855}}0.38 & 0.16                                     & {\cellcolor[rgb]{0.886,0.937,0.855}}0.22 & {\cellcolor[rgb]{0.851,0.882,0.949}}0.27 & {\cellcolor[rgb]{0.851,0.882,0.949}}0.15 & 0.18                                     & 0.13                                     & 0.19                                     & {\cellcolor[rgb]{0.886,0.937,0.855}}0.07  \\
\multicolumn{2}{c}{OpenCLIP$_{2B}$}                     & 0.78                                          & -0.07                                    & -0.37                                     & 0.10                                     & 0.44                                     & 0.08                                     & {\cellcolor[rgb]{0.851,0.882,0.949}}0.23 & 0.29                                     & 0.17                                     & 0.18                                     & {\cellcolor[rgb]{0.886,0.937,0.855}}0.11 & {\cellcolor[rgb]{0.851,0.882,0.949}}0.18 & {\cellcolor[rgb]{0.886,0.937,0.855}}0.07  \\
\multicolumn{2}{c}{OpenCLIP$_{400M}$}                   & 0.74                                          & -0.27                                    & -0.29                                     & 0.17                                     & 0.47                                     & 0.11                                     & {\cellcolor[rgb]{0.886,0.937,0.855}}0.22 & 0.33                                     & 0.18                                     & {\cellcolor[rgb]{0.886,0.937,0.855}}0.16 & 0.13                                     & 0.19                                     & {\cellcolor[rgb]{0.886,0.937,0.855}}0.07  \\
\multicolumn{2}{c}{SLIP}                            & 0.60                                          & 0.14                                     & 0.14                                      & 0.06                                     & 0.52                                     & {\cellcolor[rgb]{0.886,0.937,0.855}}0.00 & 0.24 & 0.32                                     & 0.21                                     & {\cellcolor[rgb]{0.851,0.882,0.949}}0.17 & {\cellcolor[rgb]{0.851,0.882,0.949}}0.12 & 0.19                                     & 0.09                                      \\
\multicolumn{2}{c}{DeCLIP}                          & 0.70                                          & {\cellcolor[rgb]{0.851,0.882,0.949}}0.06 & -0.17                                     & 0.11                                     & {\cellcolor[rgb]{0.851,0.882,0.949}}0.40 & 0.15                                     & 0.26                                     & 0.28                                     & 0.16                                     & 0.20                                     & 0.14                                     & {\cellcolor[rgb]{0.886,0.937,0.855}}0.17 & {\cellcolor[rgb]{0.886,0.937,0.855}}0.07  \\
\multicolumn{2}{c}{FILIP}                           & 0.45                                          & {\cellcolor[rgb]{0.851,0.882,0.949}}0.06 & 0.36                                      & {\cellcolor[rgb]{0.886,0.937,0.855}}0.01 & 0.43                                     & 0.03                                     & 0.26                                     & 0.29                                     & 0.16                                     & {\cellcolor[rgb]{0.851,0.882,0.949}}0.17 & 0.13                                     & {\cellcolor[rgb]{0.851,0.882,0.949}}0.18 & {\cellcolor[rgb]{0.886,0.937,0.855}}0.07  \\
\multicolumn{2}{c}{BLIP-2}                          & {\cellcolor[rgb]{0.886,0.937,0.855}}0.84      & -0.09                                    & {\cellcolor[rgb]{0.886,0.937,0.855}}0.07  & /                                        & /                                        & /                                        & /                                        & /                                        & /                                        & /                                        & /                                        & /                                        & /                                         \\
\multicolumn{2}{c}{GIT}                             & {\cellcolor[rgb]{0.886,0.937,0.855}}0.84      & -0.07                                    & -0.27                                     & /                                        & /                                        & /                                        & /                                        & /                                        & /                                        & /                                        & /                                        & /                                        & /                                         \\ 
\midrule
\multirow{2}{*}{\textbf{GAMA}} & \textbf{Flickr30K} & 0.81                                          & {\cellcolor[rgb]{0.851,0.882,0.949}}0.06 & {\cellcolor[rgb]{0.851,0.882,0.949}}0.09  & {\cellcolor[rgb]{0.851,0.882,0.949}}0.04 & 0.42                                     & {\cellcolor[rgb]{0.851,0.882,0.949}}0.01 & 0.26                                     & {\cellcolor[rgb]{0.886,0.937,0.855}}0.26 & {\cellcolor[rgb]{0.851,0.882,0.949}}0.15 & {\cellcolor[rgb]{0.851,0.882,0.949}}0.17 & {\cellcolor[rgb]{0.851,0.882,0.949}}0.12 & {\cellcolor[rgb]{0.851,0.882,0.949}}0.18 & {\cellcolor[rgb]{0.886,0.937,0.855}}0.07  \\
                               & \textbf{MSCOCO}    & {\cellcolor[rgb]{0.851,0.882,0.949}}0.82      & {\cellcolor[rgb]{0.886,0.937,0.855}}0.04 & {\cellcolor[rgb]{0.851,0.882,0.949}}-0.09 & {\cellcolor[rgb]{0.886,0.937,0.855}}0.01 & {\cellcolor[rgb]{0.851,0.882,0.949}}0.40 & -0.04                                    & 0.24                                     & {\cellcolor[rgb]{0.886,0.937,0.855}}0.26 & {\cellcolor[rgb]{0.886,0.937,0.855}}0.14 & 0.18                                     & {\cellcolor[rgb]{0.886,0.937,0.855}}0.11 & {\cellcolor[rgb]{0.886,0.937,0.855}}0.17 & {\cellcolor[rgb]{0.851,0.882,0.949}}0.08  \\
\bottomrule
\end{tabular}
}
\caption{\textbf{Results of zero-shot experiments on VisoGender.} The best results are highlighted in \sethlcolor{green-table}\hl{green}, and the second-best are in \sethlcolor{blue-table}\hl{blue}. The results of GAMA are obtained by the search models trained on MSCOCO and Flickr30K, respectively. Other results are retrieved from \citet{hall_visogender_2023}. Occupation-object (OO) and occupation-participant (OP) cases denote the single-person and two-person resolution tasks, respectively. For $\Delta_{RA}$ and Bias@K, closer to $0$ is better. We report mean and standard deviation ($\sigma$) for the retrieval task across all occupations.}
\label{tab:visogender}
\end{table*}

\subsection{Ablation study} \label{subsec:ablation-study}

To further explain the results, we carry out ablation studies on GAMA as well as explore the connection between object hallucination and gender bias in image captioning. Table \ref{tab:ablation-study} shows the results. We summarize the main observations as follows.

\begin{figure}[t]
    \centering
    \includegraphics[width=\linewidth]{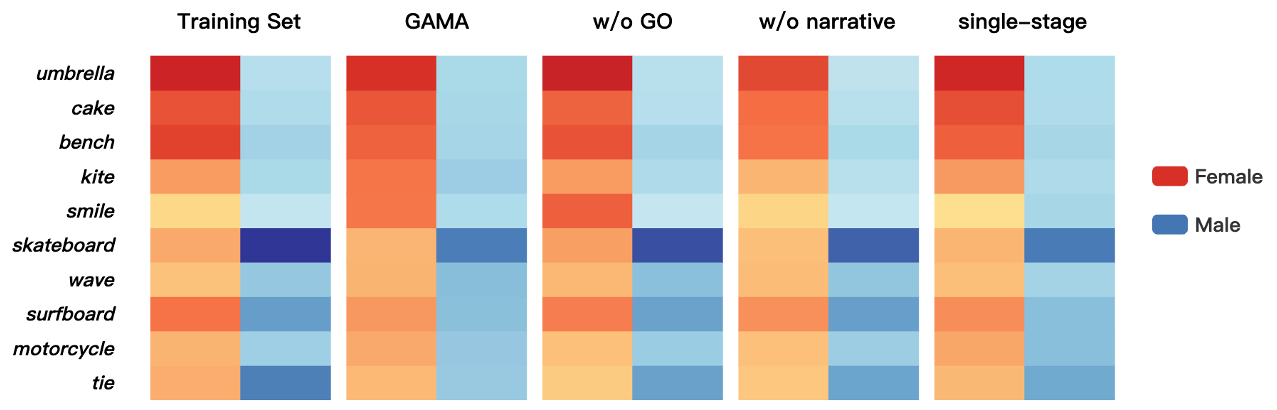}
    \caption{\textbf{Heatmap visualization of the co-occurrence frequency between gender attributes and certain words.} We respectively select five words highly co-occurring with females and males in the training set. We show the frequency of co-occurrence between gender attributes and words in the model predictions. Darker colors indicate higher frequencies.}
    \label{fig:exp-heatmap}
\end{figure}

\paragraph{The gender obfuscation module mitigates gender bias in the context.} 
First, this module effectively obfuscates gender-related information during narrative generation, leading to a reduction in the number of narratives with gender words. Next, the absence of this module is observed to cause an increase in the LIC score, indicating amplified gender bias in the generated context. Additionally, the BiasAmp score shows an increase. We speculate it is because GAMA without this module is exposed to direct gender information, leading to the generation of more words closely related to gender attributes. The rise in HR$_C^g$@10 and Figure \ref{fig:exp-heatmap} manifest that GAMA without this module hallucinates more objects that highly co-occur with gender words in the training set, thus affirming our hypothesis. Lastly, GAMA without generated narratives also achieves competitive results against previous debiasing methods. The results further demonstrate the effectiveness of the gender obfuscation module.

\paragraph{The two-stage framework can facilitate debiased answers.} 

We note an increase in the Error score in the single-stage framework. The narrative generation stage is designed to prompt GAMA to grasp the overall image before giving ultimate answers. The rise in BiasAmp and HR$_C^o$@10 indicates that the single-stage framework relies more on word co-occurrence for answer generation. Hence, the two-stage framework serves as an effective method to prevent GAMA from prematurely focusing on localized features and hallucinating ignored objects.

\paragraph{Object hallucination and gender bias exhibit a close correlation.} 
As shown in Table \ref{tab:ablation-study}, mitigating gender bias leads to a decrease in object hallucination. Object hallucination mainly results from the frequent occurrence of certain objects (including persons) and the co-occurrence between objects in the training set \citep{li_evaluating_2023}. Similarly, gender bias is manifested as the overrepresentation of a certain gender or the high-frequency co-occurrence between gender attributes and objects in datasets. In essence, gender bias can be thought of as a form of object hallucination in VLMs. Consequently, efforts to mitigate gender bias in VLMs result in a simultaneous reduction of object hallucination.

\subsection{Experiments on generalization ability} \label{subsec:generalization}

The datasets employed in the above experiments are not specifically constructed to detect gender bias in VLMs, which may contain gender bias in annotations, e.g., underrepresentation of women \citep{zhao_men_2017, harrison_run_2023}. Consequently, we turn to VisoGender for further experiments. As shown in Table \ref{tab:visogender}, GAMA exhibits good overall performance across bias metrics for both tasks, further demonstrating its generalization ability and effectiveness in gender bias mitigation.

\section{Conclusion}

In this paper, we present GAMA, a two-stage task-agnostic generation framework to mitigate gender bias towards vision-language tasks. GAMA is encouraged to gain a comprehensive understanding of images during narrative generation and to rethink gender attributes and answers during answer inference. 
Experimental results demonstrate GAMA's superiority in both task performance and gender bias metrics over previous methods. 
Furthermore, we conduct ablation studies and analyze the close connection between object hallucination and gender bias. Finally, we evaluate GAMA under a zero-shot setting to showcase its generalization ability. 
We hope that GAMA can contribute to the future exploration of fairness in VLMs.

\section*{Acknowledgements}

We would like to thank the anonymous reviewers for their insightful feedback.
This work is supported by National Key Research and Development Program of China (No. 2022YFC3302600).

\section*{Limitations}

Following previous research \citep{wang_are_2021,hirota_model-agnostic_2023}, we utilize a gender word list for pre-processing. Although prior studies have covered most gender words, some may still be omitted. One potential solution is to train a model to obfuscate gender information in a sentence via synthesized data.

Another consideration is the narrative generation model. Despite its generalization ability to various vision-language tasks and datasets, the model requires additional computing resources and datasets for training. Due to the boom in large VLMs (LVLMs), we will investigate the feasibility of generating gender-obfuscated narratives using these models.

Theoretically, we can replace the model for answer inference with any state-of-the-art task-specific generative model, which will explored in our future work.

\section*{Ethics Statement}

In our paper, we focus on mitigating gender bias towards vision-language tasks. Due to the inherent challenges associated with human analysis, including substantial manual effort and time investment, we leverage existing datasets and benchmarks for gender bias measurement. While quantitative metrics provide valuable insights, we acknowledge their potential limitations in capturing nuanced gender bias. Additionally, current datasets and benchmarks only consider binary gender, which oversimplifies the intricate and diverse nature of gender identity. Therefore, GAMA remains to be improved.
We hope that the advancements in GAMA, as presented in this paper, will serve as a catalyst for inspiring further valuable research in gender bias mitigation towards vision-language tasks.

\bibliography{citation}

\clearpage

\appendix

\section{List of Gender Words} \label{app:gender-word}

We use the gender words from \citet{hirota_model-agnostic_2023}, which are listed in Table \ref{tab:gender-word}.

\section{Metrics} \label{app:metrics}

\subsection{Gender bias metrics} 

\paragraph{LIC}

As for LIC, we follow the method proposed by \citet{hirota_quantifying_2022}.

First, we pre-process captions by masking gender words. 
To measure gender bias amplification in image captioning models, we need to quantify the difference between bias in the generated captions set $\hat{\mathcal{D}}$ and bias in the ground-truth captions in the training data $\mathcal{D}$.
Then we train two gender classifiers $f$ and $\hat{f}$ on the two masked caption sets $\mathcal{D}$ and $\hat{\mathcal{D}}$, respectively. Finally, we compare the accuracy of two gender classifiers as follows:
\begin{gather}
    \textrm{LIC}_D = \frac{1}{|\mathcal{D}|} \sum_{(y,a)\in \mathcal{D}}{s_a(y)  \mathbbm{1}[f(y)=a]}, \\
    \textrm{LIC}_M = \frac{1}{|\hat{\mathcal{D}}|} \sum_{(\hat{y},a)\in \hat{\mathcal{D}}} {\hat{s}_a(\hat{y})  \mathbbm{1}[\hat{f}(\hat{y})=a]}, \\
    \textrm{LIC} = \textrm{LIC}_M - \textrm{LIC}_D,
\end{gather}
where $y$ and $\hat{y}$ are the ground-truth caption and generated caption, respectively. We denote $a$ as the gender label of the caption. $s_a(\cdot)$ and $\hat{s}_a(\cdot)$ are the probabilities that $f$ and $\hat{f}$ classify the gender of the caption as $a$, respectively.

The higher accuracy of the classifier trained on the masked captions indicates that the context contains more information related to gender. 
$\textrm{LIC} > 0$ indicates that the model amplifies gender bias with respect to the training data, and mitigates it otherwise.

Following \citet{hirota_model-agnostic_2023}, we construct the classifiers with \texttt{bert-base-uncased} \citep{devlin_bert_2019} as the backbone and two fully connected layers with Leaky ReLU activation on top. We finetune the model for $5$ epochs with a learning rate of $1\times 10^{-5}$.

\begin{table}
\centering
\resizebox{\columnwidth}{!}{%
\begin{tabular}{cl}
\toprule
\textbf{Gender} &
  \multicolumn{1}{c}{\textbf{Gender Words}} \\ \midrule
Female &
  \begin{tabular}[c]{@{}l@{}}woman, female, lady, mother, girl, aunt, \\ 
  wife, actress, princess, waitress, sister, \\ 
  queen, chairwoman, policewoman, girlfriend, \\ pregnant, daughter, she, her, hers, herself\end{tabular} \\ \midrule
Male &
  \begin{tabular}[c]{@{}l@{}}man, male, father, gentleman, boy, uncle, \\ husband, actor, prince, waiter, son, brother, \\ guy, emperor, dude, cowboy, boyfriend, \\ chairman, policeman, he, his, him, himself\end{tabular} \\ \bottomrule
\end{tabular}%
}
\caption{Gender word list.}
\label{tab:gender-word}
\end{table}

\paragraph{Error}

Error represents the ratio of gender misclassification in the generated captions. Following \citet{hendricks_women_2018} and \citet{hirota_model-agnostic_2023}, we utilize the gender word list to identify the gender label of the generated captions as detailed in Section \ref{subsec:exp-dataset}. And ``neutral'' labels are not considered as errors.

\paragraph{BiasAmp}

BiasAmp is proposed to quantify the bias amplification of the model. Following \citet{zhao_men_2017} and \citet{hirota_model-agnostic_2023}, we use the top $1,000$ common words in captions and filter the words that are not strongly associated with humans, leaving a set $\mathcal{L}$ of high-frequency words.\footnote{The word list in available at \url{https://github.com/uclanlp/reducingbias}.}

We calculate the bias of the word $l \in \mathcal{L}$ on the gender $a\in \mathcal{A}= \{ m, f \}$ as follows:
\begin{gather}
    b_{a,l} = \frac{c_{a,l}}{\sum_{a\in \mathcal{A}}{c_{a,l}}}, \\
    \hat{b}_{a,l} = \frac{\hat{c}_{a,l}}{\sum_{a\in \mathcal{A}}{\hat{c}_{a,l}}},
\end{gather}
where $c_{a,l}$ and $\hat{c}_{a,l}$ are the number of co-occurrences of $a$ and $l$ in the training data and in the model predictions, respectively.
Then bias amplification is defined as:
\begin{equation}
    \textrm{BiasAmp} = \frac{1}{\mathcal{L}} \sum_{a\in \mathcal{A}, l\in \mathcal{L}}
    {(\hat{b}_{a,l} - b_{a,l}) \mathbbm{1}[b_{a,l} > \frac{1}{|\mathcal{A}|}]}.
\end{equation}

$\textrm{BiasAmp} > 0$ indicates that gender bias is amplified by the model, and otherwise mitigated.

\paragraph{Bias@K}

\citet{wang_are_2021} proposed Bias@K to measure gender bias in image search by comparing the proportions of masculine and feminine images in search results. Let $q$ denote the search query and $\mathcal{R}_q^K$ denote the top-K retrieved image set. The gender bias of $\mathcal{R}_q^K$ is defined as:
\begin{equation}
    \textrm{Bias@K}(q) = 
    \begin{cases}
        0,& \textrm{if}\ N_{m} + N_{f}=0 \\
        \frac{N_m-N_f}{N_m+N_f},& \text{otherwise}
    \end{cases},
\end{equation}
where $N_m$ and $N_f$ denote the number of images labeled ``male'' and ``female'' in  $\mathcal{R}_q^K$, respectively. Then Bias@K is calculated as the average of $\textrm{Bias@K}(q)$ over all queries.

A positive Bias@K indicates a higher frequency of retrieving images featuring males compared to females. This metric is most suitable when the candidate images are gender-balanced, as it does not consider the distribution of candidates.

\paragraph{MaxSkew@K}

\citet{hall_visogender_2023} measured the difference between the desired proportion of gender attributes in $\mathcal{R}_q^K$ and the actual proportion. Let $\mathcal{A}= \{ m, f\}$ denote the gender attribute set. The skew of $\mathcal{R}_q^K$ for the gender attribute $a\in \mathcal{A}$ is defined as:
\begin{equation}
    \textrm{Skew}_a\textrm{@K}(q) = \ln \frac{P_{\mathcal{R}_q^K, q, a}}{P_{d, q, a}},
\end{equation}
where $P_{d, q, a}$ and $P_{\mathcal{R}_q^K, q, a}$ are desired proportion of gender attributes in the test set and the actual proportion in $\mathcal{R}_q^K$, respectively.

Then we calculate the maximum value of Skew@K among all gender attributes for the retrieved images:
\begin{equation}
    \textrm{MaxSkew@K}(q) = \max_{a\in \mathcal{A}}{ \textrm{Skew}_a\textrm{@K}(q)},
\end{equation}

Finally, MaxSkew@K is calculated as the average of $\textrm{MaxSkew@K}(q)$ over all queries.

\paragraph{NDKL}

\citet{geyik_fairness-aware_2019} and \citet{hall_visogender_2023} measured the distance of the prediction distribution from a fair distribution over all candidate results. Let $\mathcal{R}_q$ denote the candidate set of the query $q$. The NDKL of $\mathcal{R}_q$ is defined as:
\begin{equation}
\begin{split}
    \textrm{ND}&\textrm{KL}(q) = \\ 
    &\frac{1}{Z} \sum_{K=1}^{|\mathcal{R}_q|}
    \frac{1}{\log_2{(K+1)}} d_{KL} (D_{\mathcal{R}_q^K} \| D),
\end{split}
\end{equation}
where $Z=\sum_{K=1}^{|\mathcal{R}_q|}{\frac{1}{\log_2{(K+1)}}}$ is a normalizing factor, and $d_{KL}(\cdot)$ denotes the KL-divergence. $D_{\mathcal{R}_q^K}$ and $D$ denote the actual distribution of gender attributes over the top-K retrieved images and the desired distribution, respectively. 

\paragraph{Resolution bias}

\citet{hall_visogender_2023} defined resolution accuracy (RA) as the proportion of correctly resolved pronouns. Let $\mathcal{O}$ denote the occupation set. We calculate the resolution accuracy of the gender attribute $a$ on the occupation $o \in \mathcal{O}$ as:
\begin{equation}
    RA_a(o) = \frac{\hat{n}_{a,o}}{n_{a,o}},
\end{equation}
where $n_{a,o}$ denotes the total number of pronouns of the gender attribute $a$ in occupation o, and $\hat{n}_{a,o}$ denotes the number of correctly resolved pronouns of $a$ in $o$.

Then the resolution bias on the occupation $o$ is defined as the gender resolution accuracy gap:
\begin{equation}
    \Delta_{RA}(o) = RA_m(o) - RA_f(o).
\end{equation}

Finally, we calculate the resolution bias $\Delta_{RA}$ as the average of $\Delta_{RA}(o)$ over all occupations. $\Delta_{RA} > 0$ indicates that the model performs better in resolving males within occupations, and vice versa.

\subsection{Object Hallucination Metrics} \label{app:hallucination-metrics}

\paragraph{Definition}
CHAIR \citep{rohrbach_object_2018} is a popular metric for object hallucination measurement with two variants, CHAIRi and CHAIRs. CHAIRi and CHAIRs evaluate object hallucination at the object and sentence levels, respectively. Let $N^H_o$ denote the number of hallucinated objects, $N_o$ denote the total number of mentioned objects, $N^H_c$ denote the number of captions with hallucinated objects, and $N_c$ denote the total number of captions. CHAIRi and CHAIRs are defined as:
\begin{gather}
    \textrm{CHAIRi} = \frac{N^H_o}{N_o}, \\
     \textrm{CHAIRs} = \frac{N^H_c}{N_c}.
\end{gather}

CHAIRi describes the proportion of hallucinated objects among all generated objects. CHAIRs describes the proportion of generated captions with hallucinated objects.

\paragraph{Details}

\citet{rohrbach_object_2018} utilized a synonym list \citep{lu_neural_2018} to map words to MSCOCO objects (e.g., ``player'' to ``person''). However, it is notable that the list is coarse-grained. For instance, it considers terms like ``woman'' and ``man'' as well as ``purse'' and ``briefcase'' as synonyms. Although ``woman'' and ``man'' both fall under the``person'' class, and ``purse'' and ``briefcase'' belong to the category of bags, it is crucial to recognize that they represent distinct objects. Therefore, we refine the list with hierarchical object relationships among objects based on the fine-grained classes defined in NoCaps \citep{agrawal_nocaps_2019}. If a word in a sub-category (e.g., ``woman'') is predicted in its super-category (e.g., ``person''), we do not consider it as a hallucinated object. Conversely, if a word in a category (e.g., ``woman'') is predicted in its sibling category (e.g., ``man''), we consider it as a hallucinated object.

\subsection{Hit Ratio} \label{app:HR}

\citet{li_evaluating_2023} utilized the hit ratio (HR$_C^o$) to measure the object co-occurrence in object hallucination. They demonstrated that VLMs mostly hallucinate objects that frequently co-occur with ground-truth objects in the image. 

Let $\mathcal{H}_i$ denote the set of hallucinated objects in the $i$-th image, and $\mathcal{C}_{o}$ denote the set of the top-K frequently co-occurring objects with $\hat{o}$ in the training set. The top-K hit ratio of the probing object $\hat{o}$ is defined as:
\begin{equation}
    \textrm{HR}_C^o\textrm{@K}(\hat{o}) = 
    \frac{1}{M_{\hat{o}}} \sum_{i=1}^{M_{\hat{o}}}{
    \frac{|\mathcal{H}_i \cap \mathcal{C}_{\hat{o}}|}{|\mathcal{H}_i|}
    },
\end{equation}
where $M_{\hat{o}}$ is the total number of images containing $\hat{o}$. 

Similarly, we define the top-K hit ratio of the probing gender $a$ as:
\begin{equation}
    \textrm{HR}_C^g\textrm{@K}(a) = 
    \frac{1}{M_{a}} \sum_{i=1}^{M_{a}}{
    \frac{|\mathcal{H}_i \cap \mathcal{C}_{a}|}{|\mathcal{H}_i|}
    },
\end{equation}
where $M_{a}$ is the total number of images with the gender label $a$, and $\mathcal{C}_{a}$ denotes the set of the top-K frequently co-occurring objects with the gender $a$.

\section{Implementation Details} \label{app:impl-detail}

\begin{table*}
\centering
\resizebox{0.8\linewidth}{!}{%
\begin{tabular}{ccccc} 
\toprule
\multirow{2}{*}{\textbf{Hyper-parameters}} & \multirow{2}{*}{\textbf{Narrative Generation}} & \multirow{2}{*}{\textbf{Image Captioning}} & \multicolumn{2}{c}{\textbf{\textbf{Image Search}}}                              \\ 
\cmidrule(lr){4-5}
                                           &                                                &                                            & \textbf{MSCOCO}                                           & \textbf{Flickr30K}  \\ 
\midrule
Hidden Dimension $d$                       & 768                                            & 768                                        & 768                                                       & 768                 \\
Maximum Epochs                             & 15                                             & 10                                         & 5                                                         & 5                   \\
Learning Rate                              & $4\times 10^{-5}$                                & $1\times 10^{-4}$                            & \begin{tabular}[c]{@{}c@{}}$2\times 10^{-5}$\\\end{tabular} & $3\times 10^{-5}$     \\
Weight Decay                               & 0.01                                           & 0.01                                       & 0.01                                                      & 0.01                \\
Batch Size                                 & 20                                             & 24                                         & 32                                                        & 32                  \\
Maximum Input Sequence Length              & 128                                            & 256                                        & 256                                                       & 256                 \\
Maximum Output Sequence Length             & 128                                            & 64                                         & 3                                                         & 3                   \\
\bottomrule
\end{tabular}
}
\caption{Hyper-parameter settings of GAMA.}
\label{tab:hyper-param}
\end{table*}

\begin{table*}
\centering
\resizebox{\linewidth}{!}{%
\begin{tabular}{cclc} 
\toprule
\textbf{Stage}     & \textbf{Task}        & \multicolumn{1}{c}{\textbf{Input Sequence}}                                                  & \textbf{Target Sequence}  \\ 
\midrule
1                  & Narrative Generation & Task: What can you see in this image? Please describe it in detail. Answer:~                           & {[}NARRATIVE]             \\ 
\midrule
\multirow{2}{*}{2} & Image Captioning     & Context: [NARRATIVE]. Task: Generate a short caption of the image. Answer:~                          & {[}CAPTION]               \\
                   & Image Search         & Context: [NARRATIVE]. Query: [CAPTION]. Question: Do the image and the query match? Answer:~ & Yes/No                    \\
\bottomrule
\end{tabular}
}
\caption{\textbf{The input and output formats of GAMA.} For image search, we take the token probability of the decoder as the match score. For the second stage, model-generated narratives are utilized instead of relying on ground-truth narratives from datasets.}
\label{tab:prompt}
\end{table*}

We select the best models based on the loss on the validation set for all tasks. 
Following \citet{hendricks_women_2018} and \citet{tang_mitigating_2021}, we set the beam size as $5$ during inference. Table \ref{tab:hyper-param} shows our hyper-parameters for training in different tasks. Table \ref{tab:prompt} lists the input and output formats of GAMA. 

We implement GAMA with PyTorch \citep{paszke_pytorch_2019} and Huggingface Transformers \citep{wolf_transformers_2020}. We train the models with a single NVIDIA Ge-Force RTX 4090 GPU. More implementation details can be found in our code.

\paragraph{Image search}

With a probability of $50$\%, we randomly sample the caption of another training image to create a negative pair for training. The model predicts answers with ``yes'' or ``no'' as shown in Table \ref{tab:prompt}. We take the token probability of the decoder as the match score. 


\paragraph{VisoGender} 

In VisoGender, each occupation appears with two caption templates: a person with a possessive pronoun to an object (e.g., ``the doctor and his/her stethoscope'') and a person with a possessive pronoun to a participant (e.g., ``the doctor and his/her patient''). We evaluate the performance of the trained GAMA search models on MSCOCO and Flickr30K without additional training, respectively. Our experiments on VisoGender follow the original setting proposed by \citet{hall_visogender_2023}. In the resolution task, we provide the model with queries containing ``his'' and ``her'', respectively. We determine the answers based on the match scores. In the retrieval task, we use neutral queries incorporating the pronoun ``their'' and provide the model with candidate images featuring correct occupations with balanced gender distributions.

\section{Additional Experiments}

\subsection{Experiments on VL-Bias} \label{app:vqa}

We conduct experiments on VL-Bias \citep{zhang_counterfactually_2022} under a zero-shot setting. 

\paragraph{Dataset}
VL-Bias serves as a benchmark for measuring gender bias in VLMs, encompassing $52$ activities and $13$ occupations related to humans.\footnote{The dataset is available at \url{https://github.com/VL-Bias/VL-Bias}.} The dataset includes images sourced from the Internet, as well as existing image datasets such as MSCOCO \citep{chen_microsoft_2015} and Flickr30K \citep{young_image_2014}. The captions are in the format of ``The \{gender\} is \{target\}'', where ``target'' represents an activity (e.g., ``shopping'') or an occupation (e.g., ``engineer''). In total, VL-Bias comprises $24$K image-text pairs, including $13$K pairs for the $52$ activities and $11$K for the $13$ occupations.

\paragraph{Baselines}

Following \citet{zhang_counterfactually_2022}, we evaluate GAMA against three common and effective debiasing methods: 1) Gender Swapping (\textbf{GS}) \citep{zhao_gender_2018}, which swaps gender words in the input text to mitigate the effect of gender attributes, and is adapted to vision-language tasks based on CycleGAN \citep{zhu_unpaired_2017}; 2) Dropout Regularization (\textbf{DR}) \citep{webster_measuring_2020}, which is designed to prevent model from overfitting to gender attributes by increasing the dropout rate; and 3) \textbf{FairVLP} \citep{zhang_counterfactually_2022}, which trains the PVLM ALBEF \citep{li_align_2021} with counterfactual samples to obtain debiased representations.

\paragraph{Metrics}

\citet{zhang_counterfactually_2022} proposed the vision-language bias for VL-Bias. Let $t$ denote a target word, and $a$ denote the gender attribute. The vision-language bias on $t$ towards $a$ on an image-text pair $s$ can be formulated as:
\begin{equation}
    B_{s}(t,a) = \frac
    {P_T(t|s_c) - P_T(t| s)}
    {P_A(a|s_c) - P_A(a| s)},
\end{equation}
where $s_c$ is the counterfactual image-text pair of $s$. $P_T(t|s)$ denotes the probability that the model predicts the masked target word as $t$, and $P_A(a|s)$ denotes the probability that the model predicts the masked gender word as $a$. 

Let $\mathcal{A}= \{ m, f\}$ denote the gender attribute set, and $S_t$ denote all image-text pairs with the target $t$. The gender bias over the target $t$ is defined as:
\begin{equation}
    B_{VL}(t) = \frac{1}{|S_t|}\sum_{s\in S_t}
    {B_{s}(t,m) - B_{s}(t,f)}.
\end{equation}

A positive $B_{VL}(t)$ indicates that the target $t$ is biased towards males, and vice versa. The gender bias of the dataset is calculated as the average of $B_{VL}(t)$ over all targets.

\paragraph{Implementation details}

As the baseline models are pre-trained on MSCOCO \citep{chen_microsoft_2015}, we utilize the GAMA search model trained on MSCOCO for evaluation. We take the token probability of the decoder as the probability for a target or a gender attribute. 
Counterfactual texts are constructed by reversing the gender words, while counterfactual images are generated with an adversarial attack based on the official implementation of \citet{zhang_counterfactually_2022}.

\paragraph{Experimental results}

As shown in Table \ref{tab:vl-bias}, GAMA obtains remarkable debiasing performance against baselines. The results underscore its generalization ability as well as its effectiveness in gender bias mitigation.

\begin{table}
\centering
\resizebox{0.7\columnwidth}{!}{%
\begin{tabular}{ccc} 
\toprule
\textbf{Model} & \textbf{Activity 13K}                    & \textbf{Occupation 11K}                   \\ 
\midrule
GS             & 11.21                                    & 12.47                                     \\
DR             & 11.17                                    & 13.52                                     \\
FairVLP        & 6.97                                     & 7.74                                      \\ 
\midrule
\textbf{GAMA}  & {\cellcolor[rgb]{0.886,0.937,0.855}}5.96 & {\cellcolor[rgb]{0.886,0.937,0.855}}6.83  \\
\bottomrule
\end{tabular}
}
\caption{\textbf{Results on VL-Bias.} The best results are highlighted in \sethlcolor{green-table}\hl{green}. For the results, closer to $0$ is better. The baseline results are retrieved from \citet{zhang_counterfactually_2022}.}
\label{tab:vl-bias}
\end{table}

\subsection{Further study} \label{app:ablation}

In this section, we analyze the impact of the temperature hyper-parameter, the data size of Localized Narratives \citep{voigtlaender_connecting_2023}, and the frozen parameters.

\paragraph{Temperature}

We show the results in Table \ref{tab:ablation-study-temperature}. The contrastive loss with a larger temperature is less sensitive to the hard negative samples as discussed in prior work \citep{wang_understanding_2021}. Therefore, it is hard for GAMA with a large temperature to distinguish gender-related features from gender-masked features, leading to increased LIC and BiasAmp scores.

\begin{table*}
\centering
\resizebox{0.8\textwidth}{!}{%
\begin{tabular}{ccccccccc} 
\toprule
\multirow{2}{*}{\textbf{Temperature }} & \multicolumn{3}{c}{\textbf{Gender Bias Metrics$\downarrow$}} & \multicolumn{5}{c}{\textbf{Image Captioning Metrics $\uparrow$}}                          \\ 
\cmidrule(lr){2-4}\cmidrule(lr){5-9}
                                       & \textbf{LIC} & \textbf{Error} & \textbf{BiasAmp}             & \textbf{BLEU-4} & \textbf{CIDEr} & \textbf{METEOR} & 
                                       \textbf{SPICE} & \textbf{CLIPScore}  \\ 
\midrule
0.01                                   & -1.4         & 3.5            &  -3.42                      & 37.9            & 115.2          & 30.9            & 22.5           & 75.5                \\
0.1                                    & -1.1         & 3.4            & -3.40                        & 38.2            & 115.1          & 31.0            & 22.7           & 75.4                \\
1                                      & -0.7         & 3.3            & -3.28                         & 37.7            & 114.6          & 30.6            & 22.4           & 75.6                \\
\bottomrule
\end{tabular}
}
\caption{Results of the ablation study on the temperature hyper-parameter.}
\label{tab:ablation-study-temperature}
\end{table*}

\paragraph{Data size}

Although we have demonstrated the effectiveness of the model without narrative generation in Section \ref{subsec:ablation-study}, we wonder about the influence of the data size on the results. Therefore, we conduct an ablation study on the training set size of Localized Narratives \citep{voigtlaender_connecting_2023} for narrative generation.

As our objective is to alleviate gender bias, we randomly select the images categorized under the ``Person'' class for experiments. Results are illustrated in Table \ref{tab:ablation-study-data-caption} and Table \ref{tab:ablation-study-data-viso}.

\begin{table*}
\centering
\resizebox{0.8\textwidth}{!}{%
\begin{tabular}{ccccccccc} 
\toprule
\multirow{2}{*}{\textbf{Data Size }} & \multicolumn{3}{c}{\textbf{Gender Bias Metrics~$\downarrow$}} & \multicolumn{5}{c}{\textbf{Image Captioning Metrics~$\uparrow$}}                          \\ 
\cmidrule(lr){2-4}\cmidrule(lr){5-9}
                                     & \textbf{LIC} & \textbf{Error} & \textbf{BiasAmp}              & \textbf{BLEU-4} & \textbf{CIDEr} & \textbf{METEOR} & \textbf{SPICE} & \textbf{CLIPScore}  \\ 
\midrule
100\%                                & -1.1         & 3.4            & -3.40                         & 38.2            & 115.1          & 31.0            & 22.7           & 75.4                \\ \midrule
100\% Person                         & -1.0         & 3.2            & -3.35                         & 38.0            & 114.5          & 30.9            & 22.5           & 75.3                \\
30\% Person                          & -0.8         & 3.5            & -3.02                         & 37.9            & 113.6          & 30.6            & 22.2           & 75.2                \\
\bottomrule
\end{tabular}
}
\caption{\textbf{Results of the ablation study on the data size in image captioning.} ``100\% Person'' and ``30\% Person'' denote that we randomly construct the training set for narrative generation with 100\% and 30\% of the images under the ``Person'' class, respectively. We report the average across $3$ runs.}
\label{tab:ablation-study-data-caption}
\end{table*}

\begin{table*}
\centering
\resizebox{0.9\textwidth}{!}{%
\begin{tabular}{cccccccccccccc} 
\toprule
\multirow{3}{*}{\textbf{Data Size}} & \multicolumn{3}{c}{\textbf{Resolution}}                                           & \multicolumn{10}{c}{\textbf{Retrieval}}                                                                                                                                                                                                    \\ 
\cmidrule(lr){2-4}\cmidrule(lr){5-14}
                                    & \multirow{2}{*}{\textbf{Accuracy $\uparrow$}} & \multicolumn{2}{c}{$\Delta_{RA}$} & \multicolumn{2}{c}{\textbf{Bias@5}} & \multicolumn{2}{c}{\textbf{Bias@10 }} & \multicolumn{2}{c}{\textbf{MaxSkew@5 $\downarrow$}} & \multicolumn{2}{c}{\textbf{MaxSkew@10 $\downarrow$}} & \multicolumn{2}{c}{\textbf{NDKL $\downarrow$}}  \\ 
\cmidrule(lr){3-4}\cmidrule(lr){5-6}\cmidrule(lr){7-8}\cmidrule(lr){9-12}\cmidrule(lr){13-14}
                                    &                                               & \textbf{OO} & \textbf{OP}         & \textbf{Mean} & $\sigma$            & \textbf{Mean} & $\sigma$              & \textbf{Mean} & $\sigma$                            & \textbf{Mean} & $\sigma$                             & \textbf{Mean} & $\sigma$                        \\ 
\cmidrule{1-8}\cmidrule(lr){9-10}\cmidrule{11-14}
100\%                               & 0.82                                          & 0.04        & -0.09               & 0.01          & 0.40                & -0.04         & 0.24                  & 0.26          & 0.14                                & 0.18          & 0.11                                 & 0.17          & 0.08                            \\ 
\midrule
100\% Person                        & 0.81                                          & 0.06        & 0.10                & 0.07          & 0.39                & 0.10          & 0.23                  & 0.26          & 0.16                                & 0.19          & 0.12                                 & 0.18          & 0.08                            \\
30\% Person                         & 0.76                                          & 0.14        & 0.19                & 0.11          & 0.39                & 0.15          & 0.26                  & 0.31          & 0.16                                & 0.21          & 0.13                                 & 0.19          & 0.09                            \\
\bottomrule
\end{tabular}
}
\caption{Results of the ablation study on the data size in VisoGender.}
\label{tab:ablation-study-data-viso}
\end{table*}

We observe that the data size seems to have little impact on GAMA's task performance and its gender bias mitigation ability in image captioning. However, the data size affects the generalization ability of GAMA on VisoGender. We consider that it is because a large training set enhances the zero-shot generalization ability of GAMA in narrative generation, thereby ensuring robust model performance during answer inference.

\paragraph{Parameters}

To minimize the cost of training the model, we investigate the feasibility of reducing additional training. We experiment with freezing a portion of the parameters in GAMA, and the results are summarized in Table \ref{tab:frozen-param} below.

Freezing the T5 encoder in GAMA leads to results that are only slightly inferior to those of unfrozen GAMA, which is an encouraging finding. However, the outcomes of freezing both the flan-T5 encoder and decoder are less promising. We think it is because the fusion of language features and vision features creates a new feature space, which is distinct from the original features learned by T5. Consequently, the frozen decoder cannot be effectively adapted to this new feature space.

\begin{table*}
\centering
\resizebox{0.8\linewidth}{!}{%
\begin{tabular}{ccccccccc} 
\toprule
\multirow{2}{*}{\textbf{ Model }} & \multicolumn{3}{c}{\textbf{Gender Bias Metrics~$\downarrow$}} & \multicolumn{5}{c}{\textbf{Image Captioning Metrics~$\uparrow$}}                                \\ 
\cmidrule(lr){2-4}\cmidrule(lr){5-9}
                                  & \textbf{ LIC} & \textbf{ Error} & \textbf{BiasAmp}            & \textbf{ BLEU-4} & \textbf{ CIDEr} & \textbf{ METEOR} & \textbf{ SPICE} & \textbf{ CLIPScore~}  \\ 
\midrule
GAMA                              & -1.1          & 3.4             & -3.40                       & 38.2             & 115.1           & 31.0             & 22.7            & 75.4                  \\ 
\midrule
Frozen encoder                    & -1.2          & 3.4             & -3.35                       & 38.0             & 114.6           & 30.7             & 22.2            & 75.1                  \\
Frozen encoder \& decoder           & -0.6          & 3.7             & -2.79                       & 37.1             & 112.5           & 29.8             & 21.6            & 74.4                  \\
w/o GO Frozen encoder             & 0.4           & 3.2             & -1.12                       & 37.8             & 114.2           & 30.3             & 22.1            & 74.9                  \\
\bottomrule
\end{tabular}
}
\caption{\textbf{Results of partial parameters frozen.} We freeze the T5 encoder in GAMA, both the T5 encoder and decoder in GAMA, and the T5 encoder in GAMA w/o GO, respectively. }
\label{tab:frozen-param}
\end{table*}



\end{document}